\documentclass[letterpaper]{article} % DO NOT CHANGE THIS
\usepackage{aaai2026}  % DO NOT CHANGE THIS
\usepackage{times}  % DO NOT CHANGE THIS
\usepackage{helvet}  % DO NOT CHANGE THIS
\usepackage{courier}  % DO NOT CHANGE THIS
\usepackage[hyphens]{url}  % DO NOT CHANGE THIS
\usepackage{graphicx} % DO NOT CHANGE THIS
\usepackage{natbib}  % DO NOT CHANGE THIS AND DO NOT ADD ANY OPTIONS TO IT
\usepackage{caption} % DO NOT CHANGE THIS AND DO NOT ADD ANY OPTIONS TO IT
\usepackage{algorithm}
\usepackage{algorithmic}

\usepackage{microtype}
\usepackage{booktabs}
\usepackage{lineno}
\usepackage{subfigure}
\usepackage{multirow}

\usepackage{cprotect}

\usepackage{cleveref}
\crefname{figure}{Figure}{Figures}
\crefname{table}{Table}{Tables}
\crefname{equation}{Equation}{Equations}
\crefname{section}{Section}{Sections}
\crefname{subsection}{Section}{Sections}
\crefname{algorithm}{Algorithm}{Algorithms}
\crefname{appendix}{Appendix}{Appendices}
\crefname{theorem}{Theorem}{Theorems}
\crefname{lemma}{Lemma}{Lemmas}

\usepackage{newfloat}
\usepackage{listings}
\DeclareCaptionStyle{ruled}{labelfont=normalfont,labelsep=colon,strut=off} % DO NOT CHANGE THIS
\floatstyle{ruled}
\newfloat{listing}{tb}{lst}{}
\floatname{listing}{Listing}
\title{Natural Fingerprints of Large Language Models}
\author{
    Teppei Suzuki\equalcontrib,\ Ryokan Ri\equalcontrib,\ Sho Takase\equalcontrib
}
\affiliations{
    SB Intuitions\\
    Tokyo, Japan
}

\newcommand{\avgstd}[2]{#1{\scriptsize ± #2}}
\begin{document}

\maketitle

\begin{abstract}
Recent studies have shown that the outputs from large language models (LLMs) can often reveal the identity of their source model.
While this is a natural consequence of LLMs modeling the distribution of their training data, such identifiable traces may also reflect unintended characteristics with potential implications for fairness and misuse.
In this work, we go one step further and show that even when LLMs are trained on exactly the same dataset, their outputs remain distinguishable, suggesting that training dynamics alone can leave recognizable patterns.
We refer to these unintended, distinctive characteristics as \emph{natural fingerprints}.
By systematically controlling training conditions, we show that the natural fingerprints can emerge from subtle differences in the training process, such as parameter sizes, optimization settings, and even random seeds.
These results suggest that training dynamics can systematically shape model behavior, independent of data or architecture, and should be explicitly considered in future research on transparency, reliability, and interpretability.
\end{abstract}

\section{Introduction}
\label{sec:introduction}

Recent studies have shown that large language models (LLMs) display distinctive characteristics in their outputs, enabling accurate identification of the source models~\citep{antoun2023text,sun2025idiosyncrasies,gao2024model}.
These results are largely based on experiments using LLMs trained on different datasets, where such differences naturally lead to identifiable patterns due to the models' role in modeling data distributions.
However, it remains unclear whether the distinguishability of model-generated texts arises solely from dataset variations or if other subtle factors within the training procedure itself also contribute to such characteristics.
LLMs may demonstrate skewed perspectives toward certain demographic groups~\citep{gallegos-etal-2024-bias} or produce unfair assessments when employed as evaluative tools (LLM-as-a-Judge)~\citep{zheng2023judging,chen-etal-2024-humans} due to the unintended biases.
Deeply understanding what determines the characteristics of LLM outputs is thus crucial for addressing these issues and controlling LLM behavior.

\begin{figure*}[t]
    \centering
    \includegraphics[width=0.9\linewidth]{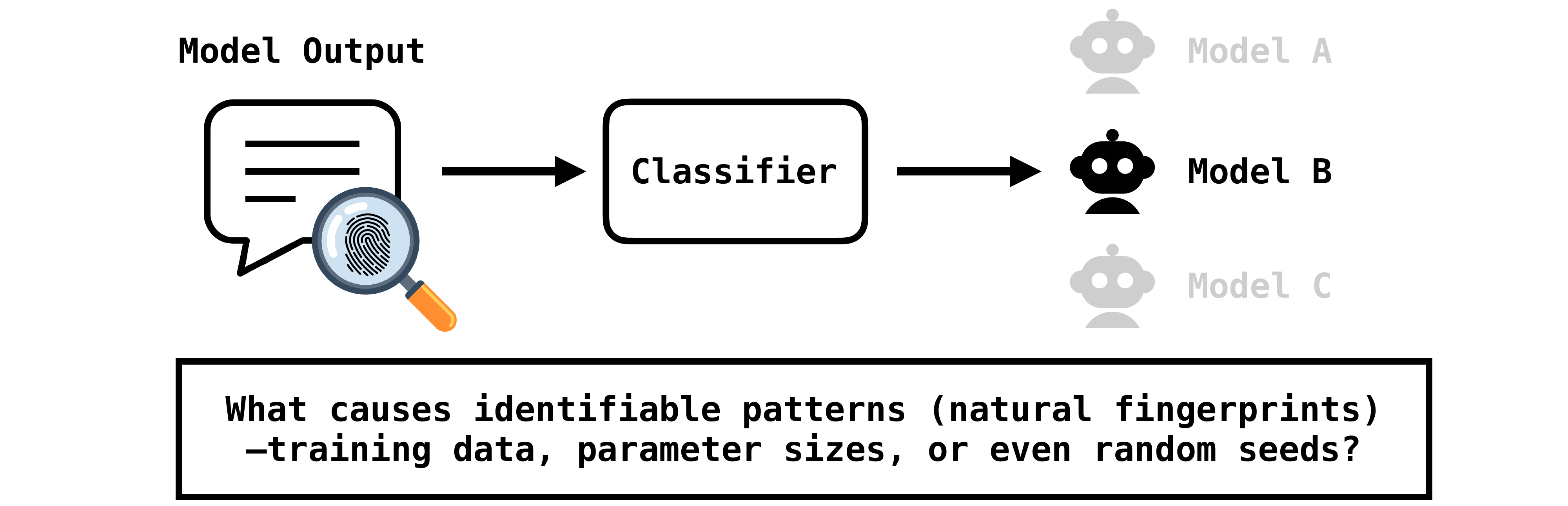}
    \caption{Overview of our experimental framework. We build a classifier to identify the source model of generated text. By varying training settings, we examine when and how ``natural fingerprints'' emerge in language models. \label{fig:overview}}
\end{figure*}

In this paper, we extend this line of research by examining subtle variations in the training process, even in scenarios where models are trained using exactly the same datasets, to determine whether these differences create identifiable model-specific characteristics.
For this purpose, we adopt dataset classification~\citep{torralba2011unbiased,liu2024decade}, which involves identifying which dataset a given data instance comes from, and high classification accuracy reveals hidden, often unintended, signatures specific to each dataset.
Analogously, we classify text samples by their source LLM, as shown in \cref{fig:overview}.
Reliable identification of model's outputs implies the existence of model-specific patterns.
This allows us to probe the underlying causes of these patterns and better understand LLM behavior.
Specifically, we begin by testing whether outputs from publicly available pretrained models can be distinguished through classification as a baseline.
We then train models from scratch under controlled settings, gradually eliminating differences in parameter size and training configurations, and examine whether the source model can still be identified from its output.
Thorough the experiments, we isolate the factors that contribute to these model-specific patterns.

Our experiments revealed several findings:
even when LLMs are trained on the same dataset, their outputs can still be distinguished above chance rate.
While it is already known that LLMs trained on different datasets produce identifiable outputs~\cite{antoun2023text,gao2024model,sun2025idiosyncrasies}, our results show that training dynamics alone can give rise to model-specific signatures.
By systematically controlling training conditions, we found that hyperparameters such as learning rate, weight decay, and even random seeds contribute to the emergence of model-specific signatures.
In particular, the training data order has a relatively strong influence on the characteristics of LLM outputs.

Unlike deliberate model fingerprinting methods~\citep{uchida2017embedding,gu2022watermarking,xu-etal-2024-instructional,yamabe2024mergeprint}, the patterns we observe emerge spontaneously during the training process without explicit design---a characteristic we term \emph{natural fingerprints}.
In doing so, our study introduces a new axis of identifiability in LLMs.
This suggests that future research on transparency, reliability, and interpretability should not only consider the influence of training data, but also account for the procedural aspects of model training.

\section{Related Work}
\label{sec:related-work}

\subsection{Bias in Large Language Models}
\label{subsec:bias-in-llms}

LLMs consistently exhibit various forms of ``bias'', referring to systematic deviations from expected or intended norms in their outputs.
These biases manifest in multiple ways: normative biases, which result in unfair treatment of certain demographic groups \citep{gallegos-etal-2024-bias}, and evaluative biases, where LLMs used as judges display systematic preferences toward particular models or content types \citep{zheng2023judging, chen-etal-2024-humans}.
Such biases are natural consequences of training data distributions or architecture's inductive biases, often unintended by developers.

Beyond these well-documented forms, we focus on more subtle structural patterns that systematically shape LLM outputs.
These subtle patterns could make generated text distinguishable, essentially serving as implicit ``fingerprints'' of specific LLMs.
In this research, we conduct classification experiments on LLM-generated texts to investigate the factors that produce these hidden biases, with broader implications for mitigating problematic biases and gaining better control over LLM behavior.

\subsection{LLM Output Identification}
\label{subsec:llm-output-identification}

Identifying LLM-generated content has become a topic of practical interest.
Distinguishing machine-generated text from human-written text is crucial for detecting the misuse of text generation models~\citep{gehrmann2019gltr,zellers2019defending,fagni2021tweepfake,solaiman2019release,guo2023close,mitrovic2023chatgpt,mitchell2023detectgpt,sadasivan2023can}.
Some research focuses on watermarking, deliberately embedding detectable traces into model outputs~\citep{pmlr-v202-kirchenbauer23a,kuditipudi2023robust,christ2023undetectable,yang2023watermarking}, or fingerprinting, identifying the model itself~\citep{uchida2017embedding,gu2022watermarking,xu-etal-2024-instructional}.
In contrast, our work addresses the fundamental question of whether model-specific traces naturally emerge during training.

Among the closest related works, \citet{antoun2023text} first demonstrated that outputs from different LLMs can be distinguished through classification.
\citet{gao2024model} proposed Model Equality Testing, a statistical framework for distinguishing black-box LLMs based on their output distributions.
\citet{sun2025idiosyncrasies} analyzed several commercial LLMs to identify distinguishing features in their outputs, finding that word choice, formatting conventions, and semantic patterns serve as recognizable characteristics.

While previous studies have examined the identifiability of LLM outputs using models trained on different datasets~\cite{antoun2023text,gao2024model,sun2025idiosyncrasies}, their experimental setups inevitably conflate the effects of data variation with other factors.
In contrast, our work isolates the impact of training configurations by holding the dataset constant and varying elements such as initialization, data order, and hyperparameters.
Our findings reveal that these seemingly minor differences can result in consistent and detectable patterns in model outputs.

\section{Experimental Design}
\label{sec:experimental-design}

\subsection{From Dataset Classification to LLM Classification}
\label{subsec:llm-classification}

Dataset classification is a task that involves identifying which dataset an image belongs to.
Originally proposed by \citet{torralba2011unbiased}, this experiment highlights unintended biases in datasets.
If classification is highly accurate, it suggests that datasets contain unique characteristics beyond their intended content, such as differences in quality, acquisition methods, or styles.

We redesign the dataset classification experiment as an LLM classification task to analyze biases in LLMs.
Formally, let $\mathcal{M} = \{M_1, M_2, \ldots, M_m\}$ be a set of $m$ distinct language models, and $\mathcal{P} = \{p_1, p_2, \ldots, p_n\}$ be a set of $n$ prompts.
For each prompt $p_i \in \mathcal{P}$ and model $M_j \in \mathcal{M}$, we obtain a generated text $t_{i,j} = M_j(p_i)$ where $M_j(p_i)$ denotes the output of model $M_j$ given prompt $p_i$.
We then construct a dataset $\mathcal{D} = \{(t_{i,j}, j) \mid i \in \{1,2,\ldots,n\}, j \in \{1,2,\ldots,m\}\}$ of pairs, where each pair consists of a generated text and the index of its source model.

Using this dataset, we train a classifier $f$ to predict the source model of a given text.
If the classification accuracy exceeds the chance rate of $\frac{1}{m}$, it indicates that each LLM embeds identifiable characteristics in its outputs, revealing model-specific biases.
By conducting this experiment with various combinations of LLMs, we aim to identify the factors that introduce biases into LLM-generated text.

In our experiments, we focus on pretrained models unless otherwise specified.
Compared to instruction-tuned models, pretrained models involve fewer design choices and a simpler construction process, making them more suitable for studying emerging natural fingerprints under controlled conditions.

\subsection{Sampling Texts from LLMs}
\label{subsec:sampling-texts}

As prompts $\mathcal{P}$, we use the first 50 characters from one million web pages sampled from the CommonCrawl subset of SlimPajama~\citep{cerebras2023slimpajama}.
Each language model $M_j$ generates continuations from the prompts using pure sampling, without adjusting decoding parameters.
The maximum number of output tokens was set to 512.
To filter out degenerate text, we removed samples where any 5-gram appeared 8 or more times.
Note that in the resulting classification dataset, the text $t_{i,j}$ is the concatenation of the prompt and the LLM's continuation.

To prevent data leakage between splits, we remove near-duplicates using semantic deduplication~\citep{abbas2023semdedup} with an English embedding model\footnote{\url{https://huggingface.co/thenlper/gte-small}}.
With the similarity threshold set at 0.2, this process removed about 0.1\% of the data.

\subsection{Classification Setup}
\label{subsec:classification-setup}

The dataset generated by LLMs is randomly split into training, validation, and test sets.
To ensure the robustness of our results, we perform the classification training and evaluation with three different random splits and report the average accuracy and standard deviation.
Both the validation and test sets contain 10,000 samples.
We experiment with two types of classifiers, each utilizing different levels of textual information.
This approach helps us determine which text features are informative for distinguishing between models.

\subsubsection{Unigram-based Classifier}
We construct logistic regression classifiers that use bag-of-subwords as features.
The accuracy of this classifier indicates whether the LLM's bias is represented in subword-level distributions.

We use LIBLINEAR~\citep{JMLR:v9:fan08a} for implementation.
We tokenize datasets with the Llama2 tokenizer~\citep{touvron2023llama2openfoundation} and use unigram frequencies to construct feature vectors.
We scale each feature value from 0 to 1.

\subsubsection{Transformer-based Classifier}
We also perform classification with a pretrained Transformer, DeBERTaV3~\citep{he2021debertav3}, with a linear classification head.
In this setup, we verify whether the text contains higher-level patterns beyond word frequencies that indicate LLM-specific biases.

For training, we use the Adam optimizer~\citep{kingma2014adam} with a batch size of 64 and a peak learning rate of 2e-5 for 1 epochs.
We employ early stopping based on accuracy on the validation set.
The learning rate is linearly warmed up in the first 10\% of the training steps and then linearly decayed to 0 over the remaining training period.

\section{Experimental Results}
\label{sec:experimental-results}

We present the results starting from comparing public LLMs trained under different settings, and progressively focus on subtle differences in training by controlling the settings where we train LLMs ourselves.

\subsection{Distinguishing Between Different LLMs}

First, we use LLMs with approximately the same number of parameters but trained on different datasets and under different conditions.
Although prior work has shown that LLMs trained on different datasets produce distinguishable outputs~\cite{antoun2023text,gao2024model,sun2025idiosyncrasies}, we begin by replicating this phenomenon in our setting.
This serves two purposes: it validates our experimental setup, and it provides a baseline against which we can compare cases where training data is held constant.

We choose six pretrained LLM families publicly available at Hugging Face Hub: Llama2~\citep{touvron2023llama2openfoundation}, Mistral~\citep{jiang2023mistral7b}, Qwen2~\citep{yang2024qwen2technicalreport}, Falcon~\citep{almazrouei2023falconseriesopenlanguage}, Pythia~\citep{biderman2023pythiasuiteanalyzinglarge}, and Sarashina2~\citep{sbintuitions2024sarashina2}.
We chose these models because their training data reportedly contains substantial amounts of English text, while still maintaining diversity (Qwen2 incorporates Chinese corpus, while Sarashina2 includes Japanese).

We use 7B-parameter pretrained models from these model families in this experiment.
We evaluate classification accuracy by varying the number of sampled texts from each LLM, ranging from 1K to 1M.

\begin{figure*}[ht]
    \centering
    \subfigure[Unigram-based classification.]{
    \includegraphics[width=0.4\linewidth]{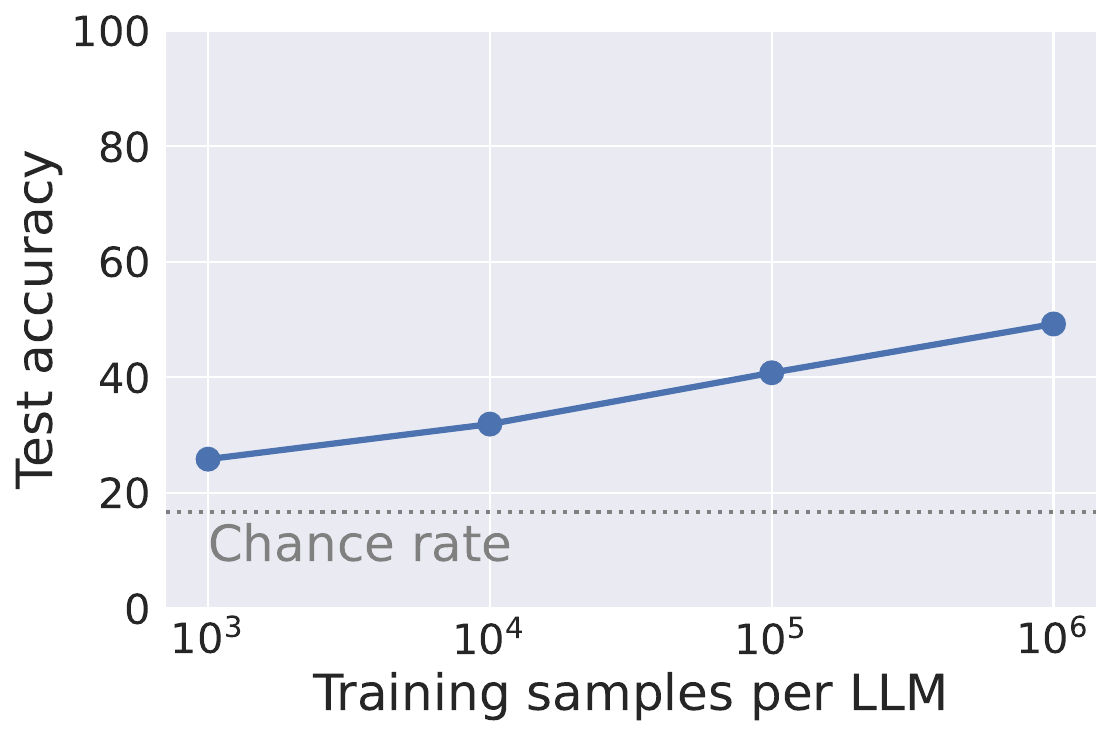}}
    \subfigure[Transformer-based classification.]{
    \includegraphics[width=0.4\linewidth]{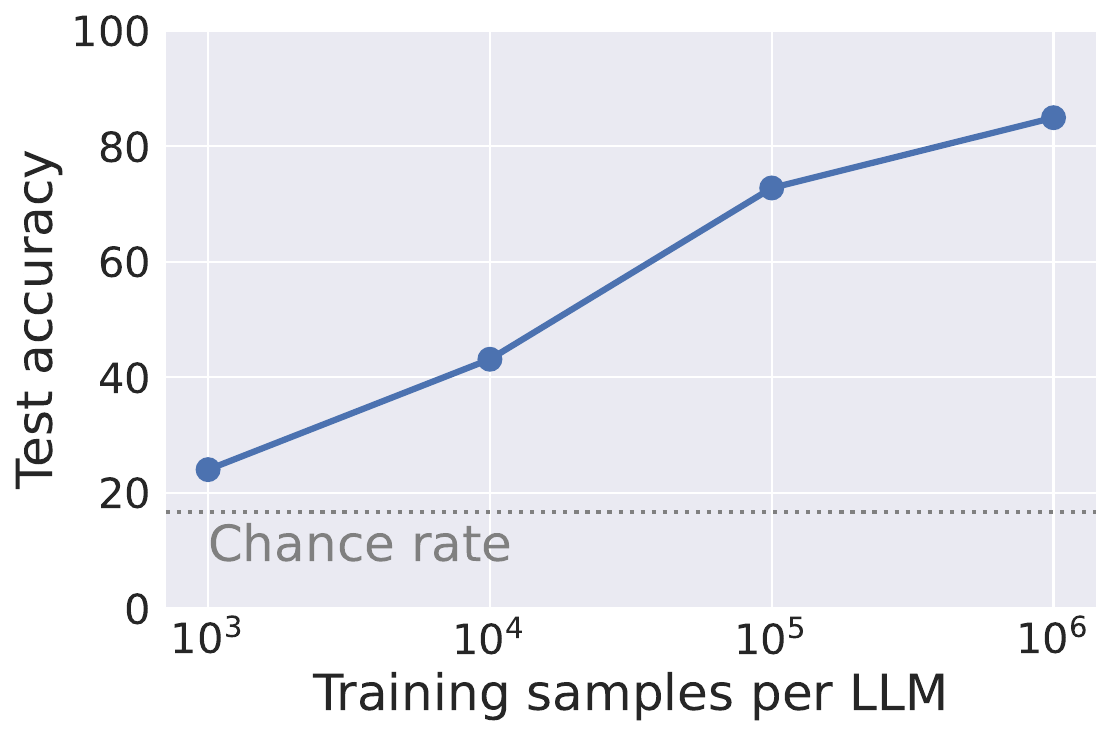}}
    \caption{Relationship between training data size and classification accuracy. The accuracy improves logarithmically as the amount of training data increases.\label{fig:7B-acc}}
\end{figure*}

\begin{table*}[!t]
    \centering
    \includegraphics[width=0.65\linewidth]{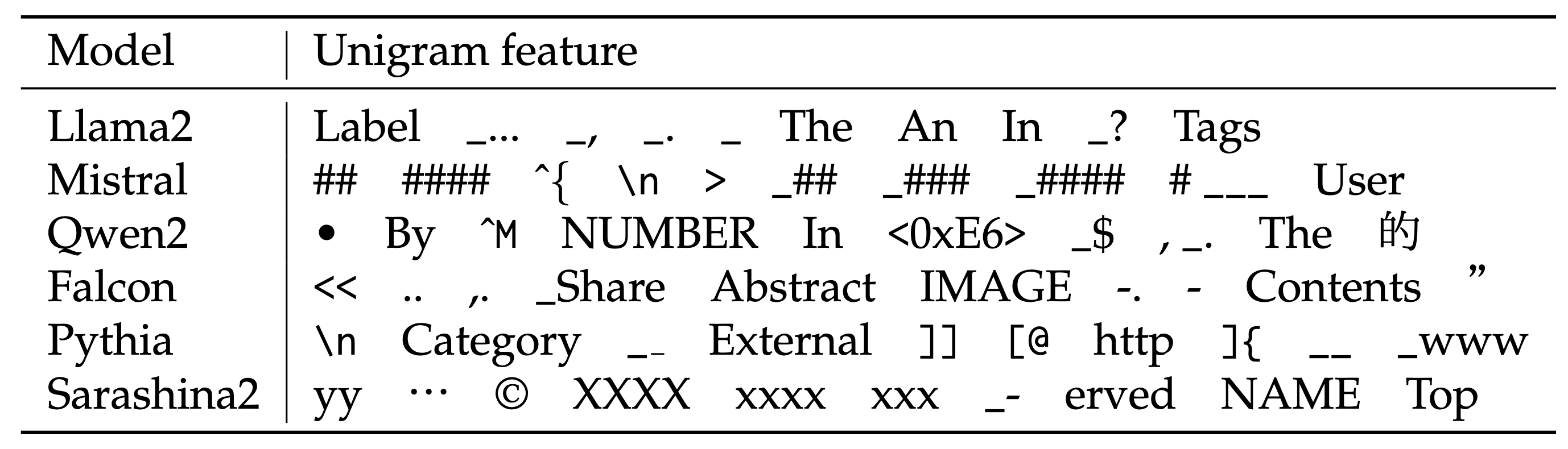}
    \caption{The top 10 largest weighted features to detect each LLM. We represent a whitespace with \textbf{\_}. \label{tab:top_features}}
\end{table*}

\cref{fig:7B-acc} shows that classification accuracy improves logarithmically as the amount of training data increases.
When using 1M training samples per LLM, the Unigram-based classifier and the Transformer-based classifier achieved accuracies of 49.2\% and 85.0\%, respectively.
These accuracies are much higher than the chance rate: 16.7\%.
This experiment revealed biases in text generated by each LLM. 
The accuracy of even simple unigram-based methods indicates bias in their unigram distributions.

\cref{tab:top_features} shows the top 10 largest weighted unigram features to detect each LLM.
Some interpretable subwords include markdown header ``\#\#'', bullet point ``•'', and tag prefixes that appear after text content such as ``Category'' and ``Tags''.
``\verb|<|0xE6\verb|>|'' from Qwen2 is a byte code that appears as part of Chinese characters.
``XXXX'' from Sarashina2 often appears as masking for personal information such as e-mail or telephone numbers.
These features likely reflect differences in training data curation and preprocessing steps used by each LLM.

While these features indicate the influence of training data on LLM bias, the models also differ in other aspects such as architecture and training hyperparameters.
Which specific differences give rise to the distinctive patterns remains under-explored, although the fact that outputs from LLMs trained on different datasets can be classified has already been established in the previous studies~\cite{antoun2023text,sun2025idiosyncrasies}.
In the following experiments, we systematically eliminate each factor by training models from scratch under controlled conditions to explore causes of bias.

\subsection{Removing the Influence of Training Data}
Based on the principles of generative models, differences in training data are expected to be a primary source of distinguishability in LLM outputs.
By examining whether outputs from different LLMs trained on the same dataset can still be distinguished, we aim to assess whether the training data itself is the main factor contributing to such distinguishability.

First, we train three models with different parameter sizes, 0.5B, 1B, and 3B, with the same training corpus from scratch.
The training corpus consists of 100B tokens extracted from FineWeb-Edu~\citep{penedo2024finewebdatasetsdecantingweb}, tokenized with the Llama2 tokenizer.
We use Megatron-LM~\citep{shoeybi2020megatronlm} for training and it takes 800 GPU hours with a H100 GPU for training a 1B model.
Other hyperparameters are described in \cref{sec:appendix-hyperparams}.

\cref{tab:different-parameters-res} shows the results from both classifiers, each trained on 1M samples per LLM.
The Unigram-based classifier achieved 39.0\% accuracy and the Transformer-based classifier reached 55.6\%.
These are lower than the accuracies observed when classifying different model families (\cref{fig:7B-acc}), despite the higher chance rate (33.3\%). 
This suggests that biases are less prominent when the training data is consistent.
This aligns with the fundamental principle that generative models learn to approximate their training data distribution.

\begin{table*}[ht]
    \centering
    \begin{tabular}{ccc} \toprule
                 & Unigram-based classification & Transformer-based classification\\ \midrule
        Accuracy (\%) & \avgstd{39.0}{0.2} & \avgstd{55.6}{0.2} \\ \bottomrule
    \end{tabular}
    \caption{Classification accuracy across 0.5B, 1B, and 3B models trained from scratch with the same training data (chance rate: 33.3\%). \label{tab:different-parameters-res}}
\end{table*}

However, the classification accuracy still exceeds chance rate, indicating that factors beyond training data differences also contribute to distinguishability.
Model size may be a major factor here, but this effect can be decomposed into two aspects: how well the model absorbs the training data distribution, and differences in the behavior of models at initialization across varying parameter spaces.
As our later experiments (\cref{sec:randomness-exp}) show, even when controlling for parameter size, initialization alone can lead to distinguishable outputs, but the accuracy decreases compared to \cref{tab:different-parameters-res}.
This suggests that the modeling capability induced by the difference in the parameter size somewhat contributes to distinguishability.

\begin{table*}[t]
    \centering
    \begin{tabular}{c c c |c c c c c c} \toprule
    \multicolumn{3}{c|}{Hyperparameter} & \\
    lr & wd & seed & PIQA & HellaSwag & WinoGrande & ARC-e & ARC-c & OBQA \\ \midrule
    5e-4 & 0.1 & 1 & 71.6 & 52.2 & 54.1 & 63.4 & 35.0 & 37.6 \\
    5e-4 & 0.1 & 2 & 70.8 & 51.4 & 54.1 & 61.7 & 34.6 & 36.0 \\
    5e-4 & 0.1 & 3 & 71.0 & 52.0 & 54.3 & 62.2 & 34.0 & 39.6 \\
    1e-3 & 0.1 & 1 & 71.5 & 53.8 & 55.5 & 64.8 & 36.4 & 38.4 \\
    1e-4 & 0.1 & 1 & 69.9 & 46.4 & 53.9 & 59.7 & 31.8 & 35.6 \\
    5e-4 & 0.01 & 1 & 70.4 & 50.5 & 52.2 & 60.7 & 33.9 & 37.4 \\
    5e-4 & 0 & 1 & 70.7 & 51.2 & 54.4 & 62.2 & 34.2 & 37.8 \\ \bottomrule
    \end{tabular}
    \caption{Zero-shot performance of our trained LLMs on common sense reasoning tasks. For hyperparameters, lr and wd mean learning rate and weight decay value, respectively.\label{tab:performance_of_trained_model}}
\end{table*}

\subsection{Removing the Influence of Parameter Sizes}
\label{sec:fixed-architecture}

Next, we evaluate whether it is possible to classify generated text when models with identical architecture (including number of layers and hidden sizes) are trained on the same dataset.
This allows us to assess the extent to which factors other than parameter size contribute to model-specific biases.

Specifically, we train seven 1B parameter LLMs with different learning rates, weight decay values, and random seeds.
We confirmed that the trained models have comparable capabilities by verifying their downstream scores are nearly identical, as shown in \cref{tab:performance_of_trained_model} which shows the zero-shot performance of these LLMs on common sense reasoning tasks: PIQA~\citep{piqa2020}, HellaSwag~\citep{zellers-etal-2019-hellaswag}, WinoGrande~\citep{10.1145/3474381}, ARC easy and challenge~\citep{clark2018thinksolvedquestionanswering}, and OpenBookQA (OBQA)~\cite{mihaylov-etal-2018-suit}.

To analyze the influence of each hyperparameter, we perform classification experiments where only one hyperparameter varies at a time.
The results in \cref{tab:diff-param-detailed} show all accuracies exceed the chance rate of 33\%.

\begin{table*}[ht]
    \centering
    \begin{tabular}{ccccc}\toprule
     \multicolumn{2}{c}{Varying Parameter} & Random Seed & Learning Rate       & Weight Decay   \\ \midrule 
     \multicolumn{2}{c}{Random Seed}   & [1, 2, 3]   &  1                  & 1              \\ 
     \multicolumn{2}{c}{Learning Rate}     & 5e-4        &  [1e-4, 5e-4, 1e-3] & 5e-4           \\ 
     \multicolumn{2}{c}{Weight Decay} & 0.1         &  0.1                & [0, 0.01, 0.1] \\ \midrule
     \multirow{2}{*}{Accuracy (\%)} & Unigram-based & \avgstd{44.5}{0.8} & \avgstd{39.4}{0.1} & \avgstd{37.9}{0.6} \\
     & Transformer-based & \avgstd{46.1}{1.6} & \avgstd{49.4}{0.1} & \avgstd{43.0}{0.3} \\ \bottomrule
    \end{tabular}
    \caption{Classification accuracy of LLMs trained with different hyperparameter settings. \label{tab:diff-param-detailed}}
\end{table*}

Compared to the previous experiment, the Transformer-based classification accuracy decreases, suggesting that differences in model size (\textit{e.g.}, particularly in the capacity to capture the training data distribution) may contribute to the emergence of higher-order distinguishable patterns beyond simple word frequency differences.

The fact that classification remains possible indicates that hyperparameter settings, such as learning rate, weight decay, or even random seed, also influence the distinguishability of LLM outputs.
This would be explained by considering how these hyperparameters guide the model toward different solutions in the loss landscape, creating subtle but detectable differences in how each model approximates the training data distribution.
The results of the Unigram-based classification show that differences in random seeds lead to the highest classification accuracy.
This suggests that differences in initialization or data ordering, or both, have a greater influence on LLM outputs in terms of word frequency distributions than changes in learning rate or weight decay.

This motivates us to further investigate the effect of randomness in a finer-grained manner.
Specifically, we examine two distinct sources of randomness: the initial weights, which determine where optimization starts, and the training data order, which influences how the optimization process proceeds.

\subsection{What kind of Randomness during Training Induces Bias in LLMs?}
\label{sec:randomness-exp}

In our training framework, the random seed affects both the model's initial weights and the order in which training data is presented.
We investigate whether we can classify models trained under three different settings:

\begin{itemize}
    \item \textbf{Order \& Init:} Both the training data order and the initial weights are randomized.
    \item \textbf{Order:} The training data order is randomized, while the initial weights remain fixed.
    \item \textbf{Init:} The initial weights are randomized, while the training data order remains fixed
\end{itemize}

The results for these settings are shown in \cref{tab:random-cm}.
All three conditions yield classification accuracies above the chance rate of 33.3\%, indicating that models remain distinguishable under each condition.
The highest accuracy is observed when both training data order and initial weights are randomized (Order \& Init).

\begin{table*}[ht]
    \centering
    \begin{tabular}{ccccc}\toprule
     \multicolumn{2}{c}{Varying Parameter} & Order \& Init  & Order  & Init \\ \midrule 
     \multirow{2}{*}{Accuracy (\%)} & Unigram-based & \avgstd{44.5}{0.8} & \avgstd{42.7}{0.6} & \avgstd{38.0}{0.4} \\
     & Transformer-based & \avgstd{46.1}{1.6}            & \avgstd{44.4}{0.6} & \avgstd{39.9}{0.3} \\ \bottomrule
    \end{tabular}
    \caption{Classification accuracy of LLMs trained with different hyperparameter settings. For Transformer-based classification, each accuracy is computed using three models trained with different hyperparameter settings. \label{tab:random-cm}}
\end{table*}

Despite training on identical data with the same objectives, simply changing the initial model weights or training data order produces distinguishable output patterns.
Notably, the Order setting yields higher accuracy in Unigram-based classification than the Init setting, suggesting that data order has a significant impact on word frequency.
In addition, the fact that the accuracy of Unigram-based classification under the Order \& Init setting is comparable to that of the Order setting suggests that initialization does not affect word frequency.
We would explain this from the practical perspective of stochastic gradient descent; early updates can have cumulative effects, or the model may be more sensitive to later updates.
These differences accumulate during training and become encoded as natural fingerprints---subtle statistical patterns in generated text that reflect each model's unique training history.
While somewhat expected in hindsight, we believe the clarity with which these effects manifest in practice is remarkable.

For insights into the characteristics of bias, we can examine the difference between Unigram-based and Transformer-based classification accuracies.
This difference is notably smaller in the randomness experiments (Table \ref{tab:random-cm}, 0.6\% to 1.7\%) compared to hyperparameter variations (Table \ref{tab:diff-param-detailed}, 1.6\% to 10.0\%). 
The reduced gap suggests that model biases induced by randomness factors are relatively subtle, primarily manifesting at the unigram distribution level rather than in the higher-order linguistic patterns captured by transformer architectures.

\begin{figure*}[ht]
   \centering
   \subfigure[Different Order Setting.]{
   \includegraphics[width=0.4\linewidth]{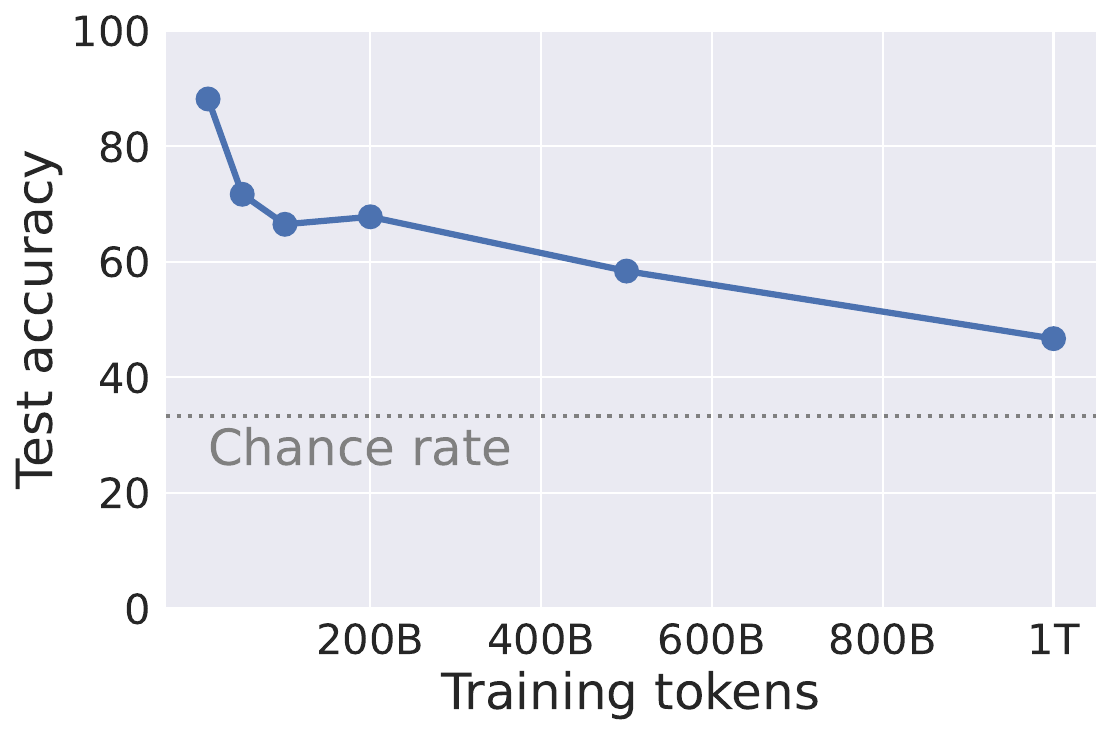}}
   \subfigure[Different Init Setting.]{
   \includegraphics[width=0.4\linewidth]{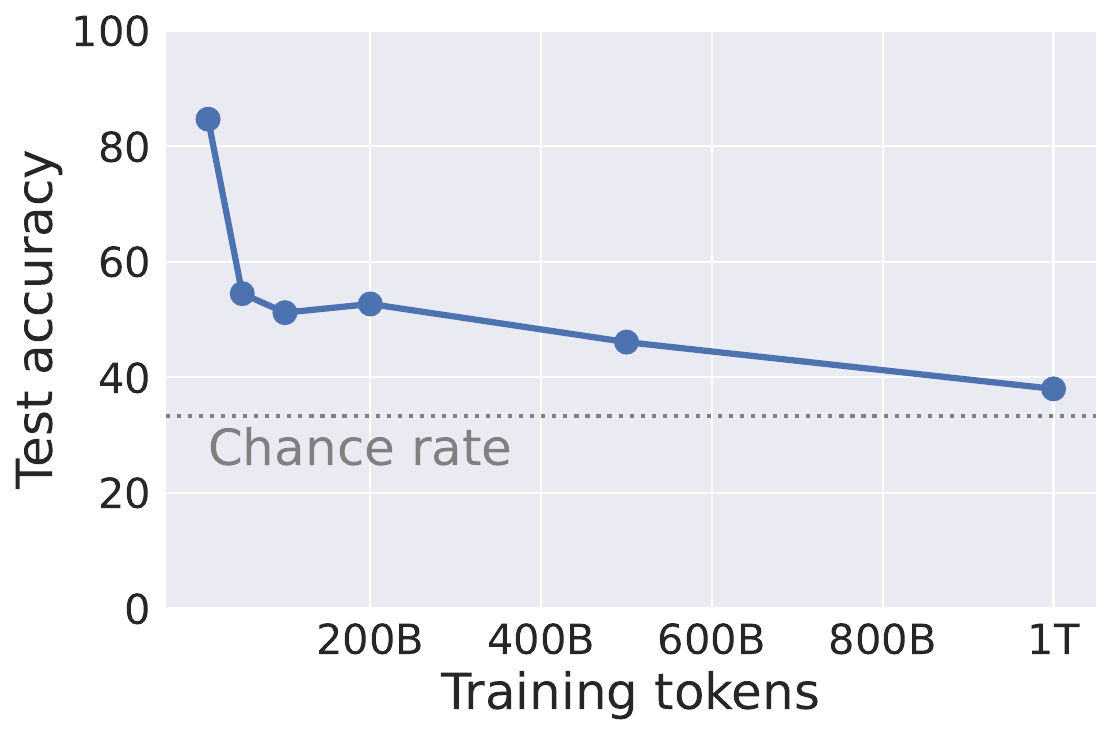}}
   \caption{Classification accuracy over the course of training for models. Figure (a) shows results for models with different training data order but the same initialization, while Figure (b) shows results for models with the same training data order but different initialization.\label{fig:accuracy-curve-in-order-setting}}
\end{figure*}

To further investigate how these ``natural fingerprints'' develop throughout the training process, we examined whether model identifiability evolves predictably over time.
We conduct additional experiments with models trained on 1T tokens instead of 100B tokens.
We train three different Transformer-based classification models under both the Order setting (different training data order, same initialization) and the Init setting (different initialization, same training data orders), and perform classification experiments at six training checkpoints: 10B, 50B, 100B, 200B, 500B, 1T tokens (\cref{fig:accuracy-curve-in-order-setting}).

In the Order setting, classification accuracy decreases over the course of training.
This can be interpreted as follows: models with identical initializations but different training data sequences initially develop distinctive patterns, but these differences diminish as training progresses and the models see the same set of training data.

Similarly, in the Init setting, models with different initializations but identical training sequences become less distinguishable over time.
This suggests that at the early stage of the training, the influence of initial model weights is prominent, but it gradually decreases as the models are exposed to the same training data.

This convergent behavior suggests that despite differences in optimization paths caused by factors such as initialization or data order, models tend to exhibit their behavior when trained sufficiently on the same underlying dataset.
However, in practical model development, it is often infeasible to perfectly control all training conditions except for the random seed and train models for sufficient steps.
In other words, completely eliminating natural fingerprints is unlikely in real-world settings.
Fortunately, the relatively small gap between accuracy of unigram-based and Transformer-based classifiers in \cref{tab:random-cm} suggests that, when training data and model size are controlled, the distinguishability may largely be attributed to differences in word frequency distributions, which in turn makes the analysis of such differences more tractable.

\subsection{Do Natural Fingerprints Emerge in Supervised Fine-Tuning?}

Finally, we test whether natural fingerprints also emerge in smaller-scale training, such as supervised fine-tuning (SFT), where models start from pretrained weights, which may have a stronger influence on training dynamics compared to the random initialization used in pretraining.

We fine-tune the 1B pretrained models (trained on 100B tokens in the previous experiment) using the Alpaca dataset~\citep{alpaca}.
We use a learning rate of 2e-5 and train for one epoch with AdamW.
Other hyperparameters are listed in \cref{sec:appendix-hyperparams}.

We experiment with two settings: Order (different training data order and same pretrained weights) and Init (different pretrained weights and same training data order).
In both settings, we train three models to make chance rate 33.3\% and run classification experiments to see if their outputs can be distinguished.

\begin{table}[ht]
    \centering
    \begin{tabular}{cccc}\toprule
     \multicolumn{2}{c}{Varying Parameter}  & Order  & Init \\ \midrule 
     \multirow{2}{*}{Accuracy (\%)} & Unigram-based & \avgstd{39.9}{0.7} & \avgstd{38.4}{0.1} \\
     & Transformer-based & \avgstd{40.3}{0.3} & \avgstd{40.8}{0.3} \\ \bottomrule
    \end{tabular}
    \caption{Classification accuracy of models fine-tuned under different conditions.\label{tab:sft-result}}
\end{table}

\begin{table*}[!t]
  \centering
  \begin{tabular}{c | l | l } \toprule
  Setting & Model & Unigram feature \\ \midrule
  %seed0
   & Model1 & uis \textbf{\_}admit \textbf{\_}== \textbf{\_}elif ". \textbf{\_}pause \textbf{\_}gentleman : eh script\\
   %seed1
   Order & Model2 & perate ah \textbf{\_}folder D 1 \textbf{\_}ac 8 \textbf{\_}Popular Focus vis\\
   %seed2
    & Model3 & \textbf{\_}Before \textbf{\_}exclus \textbf{\_}case \textbf{\_}saw \textbf{\_}Adam ]) ; \textbf{\_}bell \textbf{\_}Mike (); \\ \midrule
   %seed1
   & Model1 & \textbf{\_}Tommy ceil \textbf{\_}the scar apple HTTP \textbf{\_}Sam Over Self In\\
   %seed1111
   Init & Model2 & \textbf{\_\_} [ \textbf{\_}Forces \textbf{\_}result urchase !. ovi \textbf{\_}YES headers 0\\
   %seed42
   & Model3 & One : \textbf{\_}– Fin ’ This \textbf{\_}Louisiana ()) \textbf{\_}sep \textbf{\_}\} \\ \bottomrule
  \end{tabular}
  \caption{The top 10 largest weighted features to detect each fine-tuned LLM. We represent a whitespace with \textbf{\_}. \label{tab:top_features_of_instruction}}
\end{table*}

\Cref{tab:sft-result} shows that the Unigram-based classification and Transformer-based classification achieved 39.9\% and 40.3\% for the Order setting respectively, which means that even starting from the same pretrained model and being trained with the same data, the same models become distinguishable just by the training data order.

Also, the Init setting achieved a similar accuracies of 38.4\% and 40.8\%, indicating that even when fine-tuning on the exact same dataset in the same order, differences in the pretrained base model result in identifiable traces.
These results suggest that natural fingerprints persist even in downstream fine-tuning settings, where the scale of updates is relatively small compared to pretraining.
In particular, the fine-tuning process can amplify or preserve existing biases from pretraining, while also introducing new distinguishable patterns based on factors like data ordering.
Specifically, the data order would distort the model's output in terms of the word frequencies, as we also discussed in \cref{sec:randomness-exp}, because the accuracy of the Unigram-based classification in the Order setting is higher than that in the Init setting.

Table \ref{tab:top_features_of_instruction} shows the top 10 largest weighted unigram features used to detect each fine-tuned LLM.
Notably, many of the top unigram features appear uncommon in ordinary conversational contexts, which may reflect idiosyncrasies of the Alpaca dataset used for SFT.
Since Alpaca consists of instructions and demonstrations generated via the text-davinci-003 engine with aggressive decoding and a simplified pipeline, some tokens may originate from atypical or templated instruction formats rather than natural dialogue.
This suggests that part of the distinguishability among models may stem from how these non-natural tokens are memorized or reproduced differently depending on initialization or training dynamics.

\section{Conclusion}
In revealing that training setups alone can induce distinguishable patterns in LLM outputs, even when data and architecture are held constant, our study highlights a previously underexplored source of variability in generative models.
This phenomenon, termed natural fingerprints, highlights subtle yet consistent differences arising from minor training variations, including parameter initialization and training data order, even under identical datasets and architectures.

This suggests that future research on transparency, reliability, and interpretability should also consider training dynamics, not just dataset differences, as a source of model-specific behavior.
We hope that this work serves as a stepping stone toward a deeper understanding of the structural signals embedded in LLM outputs---and that it invites further exploration into how models subtly encode the traces of how they were trained.

\bibliography{aaai2026}

\appendix

\newpage
\onecolumn
\begin{center}
\textbf{\LARGE Appendix: Natural Fingerprints of Large Language Models}
\end{center}
\section{Hyperparameters}
\label{sec:appendix-hyperparams}

\cref{tab:hyper_params} shows that hyperparameters used in training our LLMs from scratch.
We use SwiGLU~\citep{shazeer2020gluvariantsimprovetransformer} as the activation function of FFN layers, and apply grouped query attention~\cite{ainslie-etal-2023-gqa} to self-attention layers.
We use scaled embed~\citep{takase2024spikemorestabilizingpretraining} to train some models with high learning rate: 5e-4.

\cref{tab:finetuning-hparams} shows that hyperparameters used in fine-tuning pretrained models with instruction data.

\begin{table}[!ht]
  \centering
  
  \begin{tabular}{ l | c c c } \toprule
  \textbf{Hyperparameter} & 0.5B & 1B & 3B \\ \midrule
  Layer num & 24 & 24 & 32 \\
  Hidden dim size & 1280 & 1792 & 2560 \\
  FFN dim size & 4480 & 6272 & 8960 \\
  Attention heads & \multicolumn{3}{c}{16} \\
  Dropout rate & \multicolumn{3}{c}{0.1} \\
  Sequence length & \multicolumn{3}{c}{4096} \\
  Batch size & \multicolumn{3}{c}{1024} \\
  The number of updates & \multicolumn{3}{c}{25000} \\
  Adam $\beta_1$ & \multicolumn{3}{c}{0.9} \\
  Adam $\beta_2$ & \multicolumn{3}{c}{0.95} \\
  Adam epsilon ($\epsilon$) & \multicolumn{3}{c}{1e-8} \\
  Gradient clipping & \multicolumn{3}{c}{1.0} \\
  $lr$ decay style & \multicolumn{3}{c}{cosine} \\
  $lr$ warmup step & \multicolumn{3}{c}{2500} \\
  Weight decay & \multicolumn{3}{c}{0.1} \\
  Gradient clipping & \multicolumn{3}{c}{1.0} \\
  \bottomrule
  \end{tabular}
  \caption{Hyperparameters used in training our LLMs from scratch.\label{tab:hyper_params}}
\end{table}

\begin{table}[!ht]
  \centering
  \begin{tabular}{ l | l } \toprule
  \textbf{Hyperparameter} & \textbf{Value} \\ \midrule
  Adam $\beta_1$ & 0.9 \\
  Adam $\beta_2$ & 0.95 \\
  Adam epsilon ($\epsilon$) & 1e-8 \\
  Weight decay & 0.1 \\
  $lr$ decay style & cosine \\
  $lr$ Warmup ratio & 0.002 \\
  $lr$ decay ratio & 0.9 \\
  minimum $lr$ ratio & 0.1 \\
  Sequence length & 4096 \\
  Batch size & 64 \\
  Gradient clipping & 1.0 \\ \bottomrule
  \end{tabular}
  \caption{Training hyperparameters used in our fine-tuning.\label{tab:finetuning-hparams}}
\end{table}

\end{document}